\documentclass[a4paper,fleqn]{cas-sc}

\usepackage[authoryear,longnamesfirst]{natbib}

\def\tsc#1{\csdef{#1}{\textsc{\lowercase{#1}}\xspace}}
\tsc{WGM}
\tsc{QE}
\tsc{EP}
\tsc{PMS}
\tsc{BEC}
\tsc{DE}

\usepackage{algorithm}
\usepackage{algorithmic}
\usepackage{makecell}
\usepackage{multirow}
\usepackage{mathrsfs}
\usepackage{adjustbox}
\usepackage{amsmath}
\usepackage{bbding}
\usepackage{subfigure}

\begin{document}
\let\WriteBookmarks\relax
\def\floatpagepagefraction{1}
\def\textpagefraction{.001}

\shorttitle{SelfCP: Compressing Over-Limit Prompt via the Frozen Large Language Model Itself}

\shortauthors{Jun Gao et~al.}

\title [mode = title]{SelfCP: Compressing Over-Limit Prompt via the Frozen Large Language Model Itself}                      
\tnotemark[1,2]

\tnotetext[1]{The work described in this paper was supported by the National Natural Science Foundation of China (NSFC 62106165) and Project Funded by the Priority Academic Program Development of Jiangsu Higher Education Institutions.}
\tnotetext[2]{The code and data has been released in \href{https://github.com/jungao1106/SelfCP}{https://github.com/jungao1106/SelfCP}}
\author{Jun Gao}[style=chinese]
\ead{junegao1106@gmail.com}
\credit{Conceived and designed the work, Conducted experiments, Performed the analysis, and Wrote the paper}
\affiliation{organization={Institute of Artificial Intelligence, Soochow University},
    city={Suzhou},
    country={China}}

\author{Ziqiang Cao}[style=chinese,orcid=0000-0002-1077-9033]
\cormark[1]
\ead{zqcao@suda.edu.cn}
\cortext[cor1]{Corresponding author}
\credit{Conceived and designed the analysis, Performed the analysis, and Supervised the work}

\author{Wenjie Li}[style=chinese]
\ead{cswjli@comp.polyu.edu.hk}
\credit{Conceived and designed the work, and Reviewed the work}

\begin{abstract}
Long prompt leads to huge hardware costs when using transformer-based Large Language Models (LLMs).
Unfortunately, many tasks, such as summarization, inevitably introduce long documents, and the wide application of in-context learning easily makes the prompt length explode.
This paper proposes a Self-Compressor (SelfCP), which employs the target LLM itself to compress over-limit prompts into dense vectors while keeping the allowed prompts unmodified.
Dense vectors are then projected into dense tokens via a learnable connector to make the same LLM unburden to understand.
The connector is supervised-tuned under the language modeling objective of the LLM on relatively long texts selected from publicly accessed datasets, involving an instruction dataset to make SelfCP respond to various prompts, while the target LLM keeps frozen during training.
We build the lightweight SelfCP upon 2 different backbones with merely 17M learnable parameters originating from the connector and a learnable embedding.
Evaluation on both English and Chinese benchmarks demonstrate that SelfCP effectively substitutes 12$\times$ over-limit prompts with dense tokens to reduce memory costs and booster inference throughputs, yet improving response quality.
The outstanding performance brings an efficient solution for LLMs to tackle long prompts without training LLMs from scratch.
\end{abstract}


\begin{highlights}
    \item To our knowledge, we are the first to use the frozen LLM itself to compress over-limit prompts into 1/12 memory tokens, which is more general and less expensive.
    
    \item We propose a more dialectical prompt compression perspective that achieves balances among training cost, inference efficiency, and generation quality,
\end{highlights}


\maketitle

\section{Introduction}

\begin{figure}[h]
\centering
\includegraphics[width=0.65\textwidth]{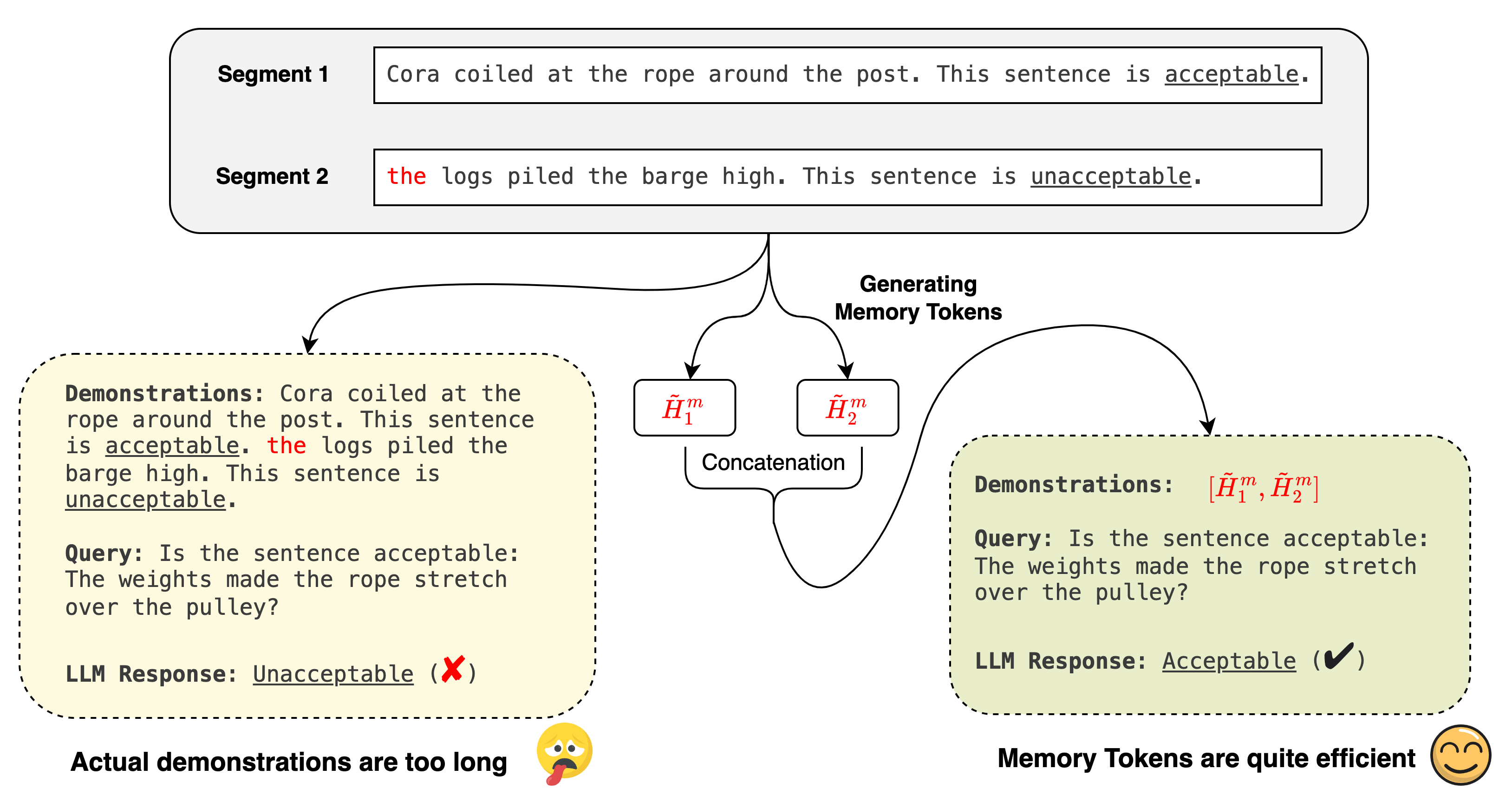}
\caption{SelfCP generates memory tokens for each segment to substitute the original in-context demonstrations, guiding the LLM to respond to the query correctly.}
\label{fig:model}
\end{figure}
Transformer-based Large Language Models (LLMs) exhibit general spectacular emergent abilities \cite{wang2023chatgpt,yang2023exploring,wei2023zero,wang12023chatgpt}.
However, the performance of LLMs heavily relies on well-designed prompts, while lengthy prompts result in memory explosion and bring other efficiency problems.
Unfortunately, prompts for many tasks have to include long inputs, such as question answering (QA) \cite{ghosal2022cicero,li2023generative}, summarization \cite{Narayan2018DontGM,sun2024dialogue,wang2024dialogue}, and query-focused summarization (QFS)~\cite{copeck2006leveraging}.
Meanwhile, the wide application of In-Context Learning (ICL) further surges the length of prompts by introducing many demonstrations.

Except for efforts in expanding the input window~\cite{zheng2022linear,wu2022memorizing,ding2023longnet,bulatov2023scaling}, previous works focus more on prompt pruning~\cite{jiang2023llmlingua} or replacing prompts with soft prompts~\cite{wingate2022prompt,ge2023context,mu2023learning}.
However, LLMs with million context windows still struggle to overcome performance degradation~\cite{liu2024lost} in addressing long prompts.
Prompt pruning methods are faced with Out-of-Distribution (OoD) problems, requiring expensive alignment between compressors and target LLMs~\cite{li2023unlocking,jiang2023llmlingua}.
Previous soft prompt compression methods respond to queries merely conditioned on soft prompts with the pursuit to minimize prompt length~\cite{ge2023context}, which increases generation difficulty and requires training the model from scratch~\cite{mu2023learning,wingate2022prompt}.
These challenges form an impossible triangle: we cannot simultaneously manage training costs, inference latency, and response quality.
Distinguished from previous attempts on soft prompt compression, we provide a dialectical perspective to compress prompts, achieving a tradeoff among these factors.

With these goals in mind, we propose SelfCP as illustrated in Figure~\ref{fig:model}, which leverages the comprehension capabilities of LLMs developed during pre-training to compress over-limit prompts.
Therefore, an extra compression module is not necessary to introduce since SelfCP employs the target LLM to compress prompts, reducing GPU memory for both training and inference initially.
Furthermore, motivated by the success of Perceiver-based Visual Language Models (VLMs) that extracted visual features via a frozen Visual Transformer (ViT)~\cite{dosovitskiy2020image}, we keep the target LLM frozen during training\footnote{For clarity, the frozen LLM plays two roles in SelfCP for compression and generation is called \textbf{compressor} and \textbf{generator} following.}, reducing the major training costs from the underlying backbone. 
However, as the built-in LLM of VLM fails to understand the visual features extracted from ViT without an adapter (e.g., Q-Former~\cite{li2023blip} or Mulit-Layer Perceptron (MLP)), a connector is important for SelfCP to convert the output features of the compressor into readable compressed tokens (called memory tokens in this paper) for the generator.
We simply utilize a linear layer supervised-tuned under the language modeling objective of the generator, serving as the connector between the compressor and the generator, and we call for more exploration for the selection of the adapter.
Previous studies responded to the query conditioned on plain soft prompts, which increased task difficulty, training costs, and wasted reasonable context window usage of their generator. 
Inspired by the fact that some segments within prompts are valuable and informative while compressing them seems not necessary\footnote{The former part of documents in text summarization are empirically more important and more informative.}, we explore an intermediate solution that allows SelfCP to only compress over-limit prompts (e.g., the originally truncated or relatively less important ones) rather than compressing everything casually.
The rest unmodified prompts are then fed into the generator directly.

This solution takes full advantage of the allowed windows wasted in the previous studies and reduces the compression and generation difficulty, but it is still faced with the following challenges: (1) The compressed soft prompt can't be posed in front always like before, since the over-limit prompts have the potential to be both the former and the latter part. (2) The over-limit prompts may be still too long for the compressor to accept while dropping directly conflicts with our goal.
Distinguished from previous studies fixing the compression mode, we diversify the compression strategies to cater to further requirements in various scenarios to approach these challenges:
(a) \textbf{Former Compression} compresses the \underline{former} half of the prompt and puts the memory tokens in front of the \underline{latter} uncompressed one.
(b) \textbf{Latter Compression} compresses the \underline{later} half of prompts and puts the memory tokens behind the \underline{former} uncompressed one.
(c) \textbf{Concatenated Compression} compress several sub-prompts (e.g., in-context demonstrations) into local memory tokens independently and concatenate them to formulate global memory tokens.
We build SelfCP upon two backbones (Vicuna-7b and BlueLM-7b) to verify the generalizability and evaluate SelfCP across of a scope out-domain tasks including generative and understanding tasks.
Then, the in-domain validation experiments show that SelfCP optimized with the three compression strategies effectively substitutes the $12\times$ over-limit prompt with soft prompts.
Our main contributions are as follows:
\begin{itemize}
    \item To our knowledge, we are the first to use the frozen LLM itself to compress over-limit prompts into 1/12 memory tokens, which is more general and less expensive.
    
    \item We propose a more dialectical prompt compression perspective that achieves balances among training cost, inference efficiency, and generation quality,
\end{itemize}

\section{Related Work}
\subsection{Soft Prompt}
Different from normal prompts consisting of discrete actual tokens, each corresponding to pre-trained embedding, soft prompts were continuous newly initialed embeddings.
On the one hand, soft prompts were usually used as a parameter-efficient method, such as Prefix-Tuning\cite{li2021prefix} and P-Tuning \cite{liu2022p}, while keeping the backbone frozen and tuning the newly initialized embeddings for each task.
On the other hand, researchers attempted to utilize soft prompts to compact prompts from concrete sequences to virtual tokens. 
Mostly from a distillation perspective, \citet{wingate2022prompt} aligned the teacher model and the student model, where the teacher model accepted the actual prompt while the student model fed the soft prompt.
The main drawback of this approach was the lack of generalization that necessitated training for each lexical different task-specific instruction.
To tackle the generalization problem, \citet{mu2023learning} proposed to learn a Llama-7b to compress instruction to virtual tokens, preceding attention past key values similar to Prefix \cite{li2021prefix}, but only compress instructions was not powerful enough since the demonstrations or input was much longer than instruction in many tasks such as summarization and QA.
To compress the long prompt, \citet{chevalier2023adapting} proposed AutoCompressor to generate compressed virtual tokens based on a fine-tuned OPT-2.7b~\cite{zhang2022opt}.
They first randomized segmented the texts with thousands of words into model-accepted range and then recursively generated soft prompts for each segment, and the previous soft prompts would be concatenated with the current segment to generate new soft prompts.
However, not only did AutoCompressor require fine-tuning on a large training set, but also encoding soft prompts would be much slower and bring extra memory cost due to Recurrent Memory Transformer (RMT) \cite{bulatov2022recurrent}.
Similarly, \citet{ge2023context} proposed ICAE that employed a LoRA-adopted Llama-7b \cite{touvron2023llama} to compress the plain processed prompts to compressed virtual tokens.
However, ICAE still struggled to address prompts longer than the allowable input window.

\subsection{Extractive Compression}
Apart from employing soft prompts, researchers also endeavored to shorten prompts by extracting informative tokens from the original ones \cite{li2023unlocking,jiang2023llmlingua}, namely token pruning \cite{kim2022learned} or token merging \cite{bolya2022token}.
Recent works like LLMLingua \cite{jiang2023llmlingua} and Selective Context \cite{li2023unlocking} shared similarities but diverged on whether to eliminate tokens with high or low Perplexity (PPL).
LLMLingua emphasized tokens with high PPL, attributing them as more influential, resulting in achieving state-of-the-art (SOTA) performance.
As mentioned in their paper extractive compression methods encountered Out-of-Distribution (OoD) issues between the extractor and the target LLM.
To reconcile this, they fine-tuned Alpaca-7b \cite{taori2023stanford} or GPT2-Alpaca using the Alpaca dataset \cite{taori2023stanford} to align to target LLMs.
However, extractive compression methods heavily hinged on the target LLM's ability to understand discrete tokens, and alignment was usually quite expensive and tailored to each target.

\subsection{Long Input Transformer}
Significant training costs and hardware support limited the input length for pre-trained language models.
A series of previous works focused on sparsifying the full attention window to linear attention \cite{dai2019transformer, child2019generating, beltagy2020longformer}, with a trade-off between efficiency and attention horizon.
Other works approximated or replaced the entire attention mechanism\cite{katharopoulos2020transformers,choromanski2020rethinking,lee2021fnet}.
However, rarefying and approximating or replacing changed the architecture of the standard transformer or training objective \cite{zhong2022training} and therefore necessitated training from scratch, which was expensive, especially scaling the model or training sets.
\cite{bertsch2023unlimiformer} proposed to offload the cross-attention computation to a single $k$-nearest-neighbor ($k$NN) index.
Although $k$NN was parameter-free, retrieving hidden states of the encoder in each generation step would slow inference, and their approaches would break down when faced with decoder-only models. 

SelfCP keeps the LLM frozen and has no limitation on building upon existing powerful LLMs.
Hence, the above approaches can further empower SelfCP theoretically.

\section{Methodology}
\begin{figure}
    \centering
    \includegraphics[width=0.95\textwidth]{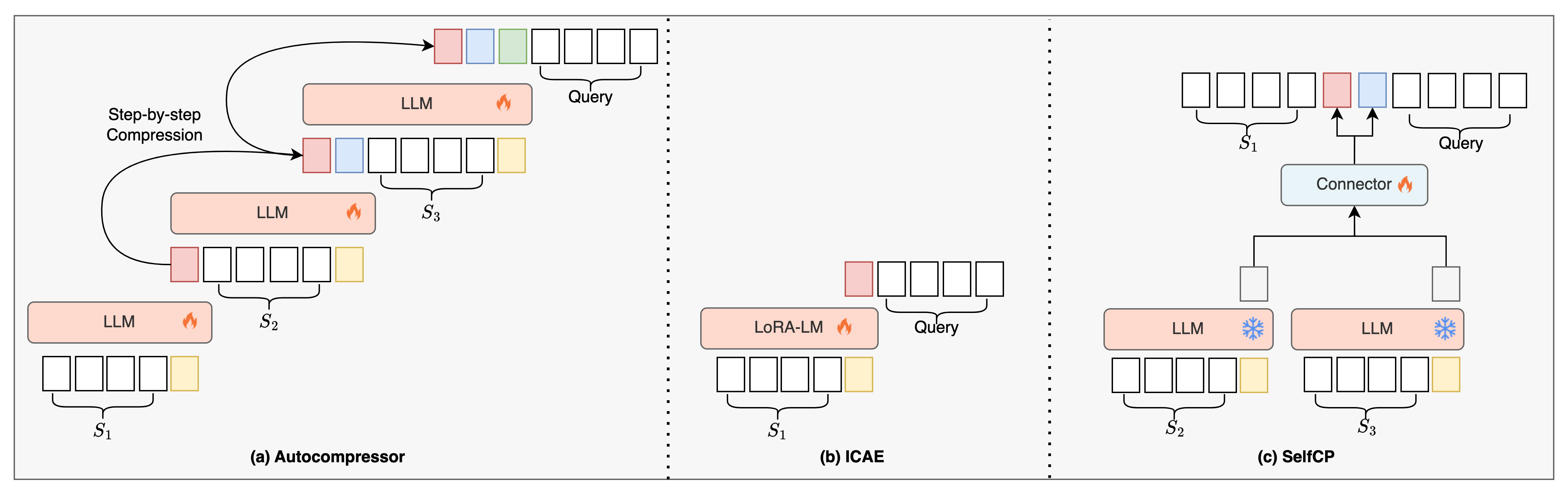}
    \caption{Differences of compression methods based on soft prompt in formulating soft prompts.
    SelfCP takes advantage of the unlimited input window in AutoCompressor and the constant compression time complexity in ICAE.}
    \label{fig:diffs}
\end{figure}
We propose the SelfCP, a parameter-efficient model that compresses prompts via the LLM itself. 
As for the selection of the underlying LLM, previous work has proved that the Decoder-only model performs better than the Encoder-Decoder model in compression instructions \cite{mu2023learning}.
We follow this conclusion and employ Vicuna-7b \cite{zheng2023judging} and BlueLM-7b~\cite{2023bluelm} as our two independent backbones, as the representation of LLMs majoring in English and Chinese.

Most compression methods based on soft prompts innovate on the compressor.
We illustrate the differences in the soft prompt formulation process for SelfCP, AutoCompressor, and ICAE in Figure~\ref{fig:diffs}.
To explain plainly, we ideally assume the compressor and the generator of three compression methods have the window limitation of $L$, have the same compression ratio, and now comes a prompt has a length $N$, where $N=3\times L$, ignoring the length of soft prompts and the query.
Considering AutoCompressor, the prompt will be divided into three segments $(S_1, S_2, S_3)$, and AutoCompressor compresses each segment step-by-step.
Notably, we provide AutoCompressor a tolerance setting that the prompt is evenly divided where they were randomly segmented originally, but AutoCompressor still requires 3 times non-parallel compression.
When it comes to ICAE, merely 1/3 prompt is accessible for the compressor and others will be read by no means.
In this case, ICAE requires only 1 time to perform compression always, since the rest is dropout.

AutoCompressor shows advantages in the readable prompt length, but is short in efficiency, while ICAE has a constant compression perplexity but struggles to approach quite long prompts.
SelfCP learns from each other skillfully that its compressor compresses each segment asynchronously and concatenates them as the AutoCompressor.
Although Selfcp compresses one less segment than AutoCompressor, the segment is directly provided to the generator to balance the overload between the compressor and the generator.
We are inclined to consider this as a trade-off among training costs, inference efficiency, and response quality.
This trade-off is more worthwhile when the unmodified segment is more important than others.

The number of practical compression steps can be calculated as $\lceil\frac{N-L}{L*k}\rceil$, where $k$ indicates that a single GPU is capable of compressing $k$ segments in a batch.
When the GPU capacity is sufficient, $k$ equals $\lceil\frac{N}{L}\rceil$, which is the scenario of ICAE that compresses all segments in a time but drops nothing, while when the GPU capacity is only sufficient to set $k=1$, it degenerates to the AutoCompressor scenario that compress segments step by step.



\subsection{Training}
The key to SelfCP is the underlying target LLM plays the role of both the compressor and the generator while keeping frozen during training.
Except for the learnable linear layer, we introduce a learnable special embedding for SelfCP, the memory tag $[M]$, initialized from a rarely used embedding of the target LLM.
To cater to future practical employment, We introduce the involved compressing strategies during training:

\subsubsection{Former \& Latter Compression}
Previous compressed methods based on soft prompts place the soft prompt at the beginning like Prefix-Tuning which ignores the basic fact that practical queries have no strict relative position relationship with corresponding prompts.
For example, we can formulate the input as \textit{Summarize the document \underline{below}: [DOC]} or \textit{[DOC] Summarize the document \underline{above}}.
Thereby, we introduce Former Compression and Later Compression instances into training.
Specifically, for prompt instances shorter than $2\times L$ tokens, namely, two times window limitation, we evenly split them into two segments $[S_c, S_u]$ and randomly compress the former or the latter.

Given the segment to be compressed $S_c$, the memory tag sequence $\mathcal{M}=[M]\times k$, and the current query $Q$, we formulate the input of compressor $I_C$ as $I_{C}=Q\oplus S_c\oplus\mathcal{M}$, where $\oplus$ represents horizon concatenation.

Memory tags signal the built-in LLM to play the role of compressor at the input end.
Then they serve as the container to absorb dense information from the preceded segment through the forward propagation of the transformer stack within the compressor.
SelfCP obtains the hidden states of the last Transformer layer on top of attached memory tags $H^m=(h^m_1, h^m_2, ..., h^m_k)$ while \textbf{disregarding others}:
\begin{equation}
    \_, H^m = Compressor(I_C).
    \label{equ:compress}
\end{equation}
Then, SelfCP projects $H^m$ sourcing from the output space of the compressor into LLM-acceptable memory tokens $\Tilde{H}^m$ via the connector, where $W_p$ is the weight of the connector:
\begin{equation}
    \Tilde{H}^m = W_p\cdot H^m.
    \label{equ:proj}
\end{equation}
Assuming the prompt has the uncompressed segment $S_u$ behind the compressed one $S_c$, the input of generator $I_G$ is formulated as $I_G=Q\oplus\Tilde{H}^m\oplus S_u$.
Given a golden response $Y=(y_1, y_2, ...,y_{|Y|})$ to the current query and prompt, the connector within SelfCP is supervised-tuned based on the language modeling objective of the target LLM in the teacher-forcing manner, where $\Theta \in \{W_p, [M]\}$:
\begin{equation}
    loss = \mathrm{maximize}_{\Theta} \prod_{i=0}^{|Y|} \mathcal{P}(y_i|I_G \oplus y_{<i}).
    \label{equ:loss}
\end{equation}

\subsubsection{Concatenated Compression}
Previous studies still struggle to process quite long prompts with efficiency or truncation problems, whose length exceeds the allowed input window of language models.
We further introduce Concatenated Compression to upgrade the previous ``Former Compression and Latter Compression'' into the training of SelfCP to tackle this problem.
Specifically, both the compressed segments $S_c$ and the uncompressed ones will exceed $L$ tokens when prompts longer than $2\times L$ tokens after even partition.
In this scenario, given a prompt has $N$ tokens in total, where $N>2\times L$, SelfCP first allocates uncompressed segments $S_u$ with $L$ tokens, and then evenly splits the rest into $\lceil \frac{N}{L}-1 \rceil$ non-overlapping local segments.
Due to the segments being non-overlapping, the compressor compresses each of them independently as Equ.~\ref{equ:compress} and converts the hidden states on top of memory tags to local memory tokens as Equ.~\ref{equ:proj}.
The global memory tokens are configured by concatenating local memory tokens horizontally, fed into the generator to optimize SelfCP as Equ.~\ref{equ:loss}.

\subsection{Improvement for In-Context Learning}
Specifically, in the scenario of ICL, we consider the prompt containing the demonstrations and the query containing related task inputs and task instructions.
In-context demonstration sequence typically increases the length of the prompt, we specially develop strategies to optimize both the efficiency and effectiveness of ICL through caching and retrieving.

SelfCP allocates each demonstration with a segment and truncates the latter to guarantee independence among each segment.
Hence, SelfCP compresses each demonstration independently and caches its memory tokens to construct a Memory Demonstrations Bank (MDB) for reusing.
With the help of MDB, SelfCP can respond to queries by directly withdrawing target memory tokens without repetitive compression, making SelfCP more efficient in ICL.

Randomly sampled demonstrations have unstable ICL performance gains.
We further empower MDB to support demonstration selection by treating the first memory token $k_i$ of $i-$th demonstration as the key used for retrieval.
SelfCP requires compressing the query $Q$ to obtain the first memory token $q$ for directional demonstration selection.
Then, the demonstration scores of the $i$-th demonstration in MDB are calculated based on cosine similarity:
\begin{equation}
    score_i = cosine(q, k_i).
\end{equation}
SelfCP selects the target demonstrations according to the demonstration scores from high to low, being more effective in ICL.

\section{Experiment}

\subsection{Datasets}
\begin{figure}
    \centering
    \includegraphics[width=0.6\textwidth]{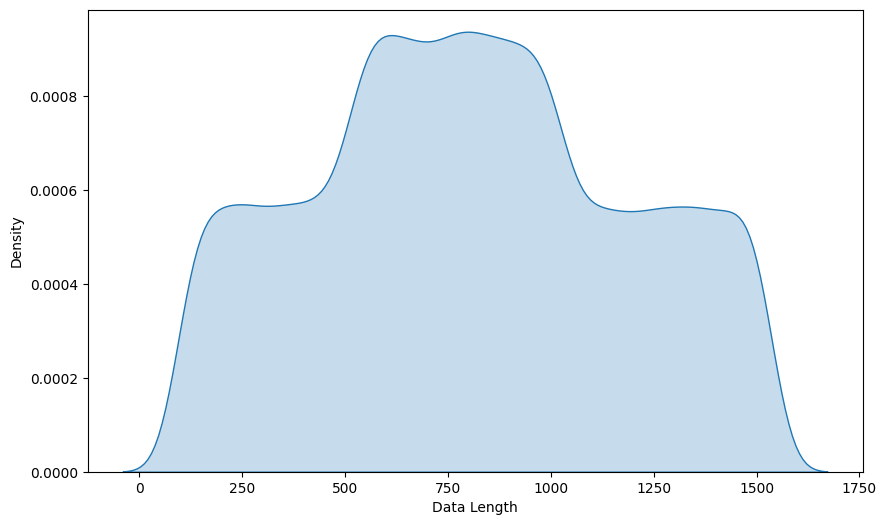}
    \caption{Kernel density estimation of training data. The horizontal axis is the number of tokens in each instance.}
    \label{fig:dense}
\end{figure}
\begin{table}
  \centering
  \caption{The experiment details of test sets used in this paper.  ICL represents performing ICL or not. LA stands for linguistic acceptability.}
    \begin{tabular}{lcccc}
    \toprule
    Task  & Dataset  & In-Domain & ICL &\# Test Instances \\
    \midrule
    SUM & XSUM   & \Checkmark & \XSolidBrush&11,334\\
    \midrule
    QA   & CICERO   & \Checkmark & \XSolidBrush & 10,898\\
    \midrule
    QFS & DUC     &  \XSolidBrush & \XSolidBrush & 45  \\
    \midrule
    CVG & CLCV    &  \XSolidBrush & \XSolidBrush &2,000  \\
    \midrule
    SUM & XSUM    & \Checkmark & \Checkmark & 1,500\\
    \midrule
    SUM & ARXIV    & \XSolidBrush & \Checkmark & 1,500\\
    \midrule
    LA & CoLA & \XSolidBrush &\Checkmark & 1,041\\
    \bottomrule
    \end{tabular}%

  \label{tab:settings}%
\end{table}%

We mix up XSum, CICERO, and an instruction dataset as our training set, containing 42k instances, and the instance length kernel estimation is illustrated in Figure~\ref{fig:dense}.
The details of in- and out-domain test sets are shown in Table \ref{tab:settings}. 
Notably, we don't perform in-domain evaluation in the instruction dataset as its responses are somewhat open-ended and it solely has a training set.
We use the entire test set of XSUM and CICERO as the in-domain set to confirm details of prompt compression.
In out-domain evaluation, we employ the entire DUC 2007 as the test set, and we collect a Chinese verdict generation (CVG) dataset called CLCV (\textbf{C}hinese \textbf{L}egal \textbf{C}ase \textbf{V}erdict).
CLCV has 2,000 instances collected from China Prosecutorial Network\footnote{\url{https://www.12309.gov.cn/12309/zjxflws/index.shtml.}}.
Each instance of CLCV contains the indictment and the corresponding judgment with an average of 540 words for indictment and 230 words for judgment.
When evaluating the ICL performance of SelfCP, we sampled 1,500 instances from the XSUM and ARXIV summarization datasets respectively, with an average length of about 580 words.
Additionally, we evaluate SelfCP on the entire linguistic acceptability dataset CoLA through close-ended evaluation in ICL.

\subsection{Evaluation Metrics}
ROUGE \cite{lin2004rouge} is a widely adopted metric in many generative tasks that evaluate how similar the generated hypothesis is to the golden label.
Therefore, ROUGE is used in our experiments to evaluate the quality responses generated conditioned on compressed virtual tokens.
We report the F-1 scores of ROUGE-1, ROUGE-2, and ROUGE-L (abbreviated R-1, R-2, R-L in the following), and we employed the files2rouge \footnote{\url{https://github.com/pltrdy/files2rouge.}} library in practice.
Additionally, as CLCV is a Chinese dataset, we use the Chinese-ROUGE \footnote{\url{https://pypi.org/project/rouge-chinese.}} library combined with jieba word-cutting library \footnote{\url{https://pypi.org/project/jieba.}} to evaluate the generated result.
For CoLA, we report the accuracy.

\subsection{Baselines}
We compare the performance of SelfCP with the naive LLMs fed actual prompts based on Vicuna-7b and BlueLM-7b, respectively.
In addition, we introduce AutoCompressor~\cite{chevalier2023adapting} and ICAE~\cite{ge2023context} which converts the entire prompts into virtual tokens, and LLMlingua~\cite{jiang2023llmlingua} which drops uninformative tokens in the prompt. 

\paragraph{AutoCompressor}
AutoCompressor\footnote{ \url{https://github.com/princeton-nlp/AutoCompressors.}} is the recent work that compresses prompts into virtual tokens recurrently with the fine-tuned LLMs \cite{zhang2022opt}.
We employ their officially released weight based on Llama2-7b and compare its performance with SelfCP on out-domain datasets, setting the compression ratio the same as SelfCP.

\paragraph{ICAE}
ICAE\footnote{\url{https://github.com/getao/icae.}} compresses entire prompts with given window limitation by a LoRA-adapted Llama2.
We employ their officially released Llama2-7b version and compared its performance with SelfCP in out-domain datasets, setting the compression ratio the same as SelfCP.

\paragraph{LLMLingua} 
LLMLingua is a recent coarse-to-fine prompt compression method based on dropping uninformative words and achieves powerful performance.
We employ LLMLingua from their official code\footnote{\url{https://aka.ms/LLMLingua.}}, and compare its performance with SelfCP in all out-domain datasets, setting the compression ratio the same as SelfCP by limiting the number of dropped discrete tokens.
Notably, LLMLingua, in their paper, is designed to employ a small compressor (Llama or GPT-2), instruct-tuned to align with the target LLM (GPT-3.5-Turbo or Claude-v1.3).
For a meaningful comparison, we substitute their target LLMs with the underlying LLM in SelfCP.

\subsection{Settings}
Considering the max tokens in all involved datasets and computation efficiency, we set the max allowed input window limitation $L$ to 512.
Additionally, we fix the learning rate to 8e-5 and use Adam as the optimizer, and the effective batch size is 32 (8 GPUs data parallelism and 4 steps gradient accumulation).
Additionally, we conduct all experiments on 8*NVIDIA A5000 24G GPUs based on BFloat 16 data type.

We compress the \textbf{latter} part in XSUM, DUC, and CICERO, since the former part in these tasks is empirically important, while we compress the \textbf{former} part in CLCV because the involved person is introduced at the front of the indictment which is relatively unimportant.
Additionally, we divide ICL into two situations:
(1) In the low-resource situation, we fix the demonstrations for each query.
(2) In the high-resource situation, SelfCP retrieves similar demonstrations from MDB by measuring the cosine similarity.
We consider the training set as the demonstration pool and construct the MDB for each dataset.

\subsection{Results}

\subsubsection{In-Domain Evaluation}
\begin{table}
\centering
\caption{The in-domain results of XSUM and CICERO. 1k represents extending window limitation to 1k, and the others is 512.
}
\begin{tabular}{llccc|ccc}
\bottomrule
\multirow{2}{*}{Backbone}& \multirow{2}{*}{Method} & \multicolumn{3}{c|}{XSUM} & \multicolumn{3}{c}{CICERO} \\
&&R-1&R-2&R-L&R-1&R-2&R-L\\
\hline
\multirow{5}{*}{Vicuna-7b}&Vicuna&19.9&5.0&13.5&17.3&3.3&14.3\\
& \quad\textit{\footnotesize+LoRA} & 25.4 & 7.5 & 17.3 & 28.1 & 10.5 & 25.6\\

&Vicuna-1k&27.3&8.7&19.7&30.5 & 11.3&27.4\\
& \quad\textit{\footnotesize+LoRA}  & 31.2 & 11.0 & 23.1 & 34.1 & 13.5 & 30.2 \\

&SelfCP&30.5 &{10.8}&22.7&{33.3}&{12.9}&{29.2}\\
\hline
\multirow{5}{*}{BlueLM-7b}&BlueLM &15.0&3.6&10.4&17.6&3.1&15.0\\
& \quad\textit{\footnotesize+LoRA} & 23.1 & 7.6 & 17.4 & 21.9 & 7.8 & 19.8\\
&BlueLM-1k &28.1& 9.9&22.8&25.1&9.2&23.1\\
& \quad\textit{\footnotesize+LoRA} & 30.8 & 10.5 & 24.6 & 31.2 & 10.8 & 27.4\\
&SelfCP&{30.9} &10.6&{24.4}&29.9&10.6&26.9\\
\toprule

\end{tabular}

\label{tab:in-domain}
\end{table}
We conduct the in-domain evaluation on XSUM and CICERO in Table \ref{tab:in-domain}.
SelfCP significantly outperforms the baseline Vicuna and BlueLM with 512 and 1024 window limitations after supervised tuning.
To simulate gains brought by the trained connector, we also LoRA-tune Vicuna and BlueLM on our training set with 17M trainable parameters by setting the LoRA rank to 32 (referring to $+LoRA$ in Tab.~\ref{tab:in-domain}).
In this case, SelfCP outperforms LoRA-adapted backbones with 512 allowed windows even the model being tuned, while SelfCP is comparable with them with 1,024 allowed windows.
These results highlight that extreme truncation makes LLMs confused to respond and the compressed virtual tokens effectively filter noise information and recover the informative parts to a large extent.

\subsubsection{Out-Domain Evaluation}
\begin{table}[t]
\centering
 \caption{The out-domain results of DUC and CICV.}
\begin{tabular}{llccc|ccc}
\bottomrule
\multirow{2}{*}{Backbone}&\multirow{2}{*}{Method}  &\multicolumn{3}{c|}{DUC} &\multicolumn{3}{c}{CLCV} \\
&&R-1&R-2&R-L&R-1&R-2&R-L\\
\hline
\multirow{2}{*}{Llama2-7b} & AutoCompressor  & 31.7 &7.3&16.4 &20.8  &6.1   &17.0\\
 & ICAE  & 30.6 &6.9 &15.7 &22.3  &6.4   &19.2\\
\hline
\multirow{3}{*}{Vicuna-7b}&Vicuna &24.0& 4.6&13.1&20.6    &6.0   &16.5\\
&LLMLingua &32.5&7.6&16.8&21.0 &6.3 & 17.2\\
&SelfCP &{33.6}&{7.8}&{17.3}&21.4    &6.3    &18.0\\
\hline
\multirow{3}{*}{BlueLM-7b}&BlueLM &16.8& 3.6&10.0&33.9    &17.5   &24.8\\
&LLMLingua &21.2&4.1&11.4 &35.1 &18.6 &25.9 \\
&SelfCP &25.5& 4.8&13.4&{36.6}    &{20.8}    &{27.2}\\
\toprule
\end{tabular}

\label{tab:QFS}
\end{table}

We test the out-domain performance of SelfCP on the DUC and CLCV to evaluate its generalized capability and cross-lingua ability, as demonstrated in Table~\ref{tab:QFS}.

SelfCP employs concatenation compression to compress the query-related documents into memory tokens.
Compared with the truncation that occurs in naive Vicuna and BlueLM, SelfCP effectively broadens the in-context window, achieving nearly +10 ROUGE-1 gains.
While in CLCV, BlueLM-based SelfCP achieves better performance compared with Vicuna-based ones since BlueLM is specific-tuned and good at Chinese, proving that SelfCP implicitly leverages the strengths of diverse backbones during learning prompt compression.
Additionally, the compression on both the former of CLCV and the latter part of DUC indicates that SelfCP makes no limitation on the position of memory tokens.
As for AutoCompressor,
the 7b version underperforms Vicuna-based SelfCP in English tasks (DUC) and underperformes BlueLM-based SelfCP in Chinese tasks (CLCV).
Meanwhile, it isn't surprising to find that SelfCP outperforms LLMLingua in out-domain evaluation since their algorithm leverages the understanding ability of the target LLM while ChatGPT-3.5-turbo is much stronger than LLM with 7b parameters.
Therefore, Vicuna- or BlueLM-7b may sometimes be confused about the meaningless discreet tokens.

\subsubsection{In-Context Learning Evaluation}
\begin{table}[t]
\centering
\caption{The in-context learning results in XSUM and ARXIV of SelfCP based on Vicuna.
$\dag$ represents the demonstrations retrieved from MDB, and others uses fixed demonstrations.}
\begin{tabular}{lc|ccc|ccc|c}
\bottomrule
\multirow{2}{*}{Method} & \multirow{2}{*}{\#-shots} &\multicolumn{3}{c|}{XSUM} &\multicolumn{3}{c|}{ARXIV} & CoLA \\
&&R-1&R-2&R-L &R-1&R-2&R-L & Acc.\\
\hline
\multirow{2}{*}{Vicuna} &0-shot &19.9& 5.0&13.4 &34.3& 9.1&27.4 & 56.2\\
 &1-shot &21.2& 5.8&14.5&34.4& 9.1&27.5 & 57.4\\

\hline
\multirow{3}{*}{AutoCompressor} &1-shot & 20.3 &6.3 &13.7 & 26.4	&8.2&25.8 & 40.9\\
 &2-shot & 21.4 & 6.4 &14.1&26.2 &7.9&25.4 & 56.3\\
 &5-shot & 21.7 & 6.7 &14.3&33.7 &9.1&27.9 & 58.8\\
 \hline

\multirow{3}{*}{ICAE} &1-shot &21.9 &7.8 &20.3 
& 24.6 & 7.1 &22.9 & 30.9\\
 &2-shot &23.2 &8.4 &20.8 & 25.5& 7.6 &24.3 & 30.9\\
  &5-shot &24.9 &8.8 &21.6 & 26.9 & 7.7 &25.8 & 30.9\\
\hline

\multirow{3}{*}{LLMLingua} &1-shot &19.7 &5.2 &14.4 & 33.1 & 8.7 &27.1 & 55.0\\
 &2-shot &20.0 &5.1 &14.1 & 32.0 & 8.1 &25.9 & 55.7\\
  &5-shot &18.6 &4.9 &14.3 & 29.7 & 7.4 &24.6 & 56.9\\
\hline 

\multirow{3}{*}{SelfCP} 
&1-shot  &25.5 &9.1    &20.0&34.7 &10.2 & 27.9 & 58.0\\
 &2-shot &{26.8}  &9.8   &20.9&35.2	&10.4 &28.2 & 61.0\\
 &5-shot &{27.6}  &10.1   &21.4&35.4	&10.2 &28.1 & 61.8\\

\hline
\multirow{3}{*}{SelfCP$^\dag$} 
&1-shot &26.2&9.4&20.7&34.7&{10.3}&27.8 & 58.7\\
& 2-shot &{28.9}&{10.5}&{21.7}&{34.3}&{10.4}&{28.3} & 61.2\\
& 5-shot &{30.0}&{11.2}&{22.3}&{35.3}&{10.8}&{27.7} & 61.5\\
\toprule
\end{tabular}

\label{tab:context-xsum}
\end{table}

\begin{table}[t]
\centering
\caption{The in-context learning results of SelfCP based on BlueLM.}
\begin{tabular}{lc|ccc|ccc|c}
\bottomrule
\multirow{2}{*}{Method} & \multirow{2}{*}{\#-shots}  &\multicolumn{3}{c|}{XSUM} &\multicolumn{3}{c|}{ARXIV} & CoLA \\
&&R-1&R-2&R-L &R-1&R-2&R-L & Acc.\\
\hline
\multirow{2}{*}{BlueLM} & 0-shot &15.0& 3.6&10.4 &30.9& 7.7&24.7 & 71.6 \\

 & 1-shot &19.1& 4.8&12.1&23.0& 3.6&19.0 & 69.6\\
\hline 
\multirow{3}{*}{LLMLingua} & 1-shot &18.0 & 2.7 & 13.2 &28.0 & 6.3 & 23.2 & 58.1 \\
 & 2-shot &18.6 & 3.4 & 13.4 &26.5 & 5.5 & 22.2 & 60.3\\
 & 5-shot &18.3 & 3.3 & 13.3 &26.8 & 5.6 & 22.4 & 62.5 \\
\hline
\multirow{3}{*}{SelfCP} 
 & 1-shot &24.0 & 6.9  &18.0 &31.4 &7.7 & 25.2 & 69.6\\
 & 2-shot &25.0  &7.3   &18.8 & 30.8 & 7.3 &24.8 & 70.1\\
 & 5-shot &25.3  &7.4   &19.1 & 31.9 & 7.8 &26.0 & 71.8\\
\hline
\multirow{3}{*}{SelfCP$^\dag$} 
& 1-shot &24.7 &7.2&18.5&31.0&7.5&24.9 & 70.1\\
& 2-shot &{25.1}&{7.4}&{19.0}&{31.2}&{7.7}&{25.1} & 70.3\\
& 5-shot &{26.3}&{7.6}&{20.0}&{31.5}&{7.9}&{25.3} & 71.1\\
\toprule
\end{tabular}
\label{tab:context-arxiv}
\end{table}

We evaluate the ICL ability of SelfCP in Table \ref{tab:context-xsum} and \ref{tab:context-arxiv}.
Memory tokens take equal or even better effects than actual demonstrations, which verifies the ability of SelfCP to capture the core semantics of demonstrations during compression.
Regarding ARXIV, the original ICL is not helpful enough and causes significant degradation in BlueLM, due to the relatively long documents in ARXIV, which leave little room for LLM to read demonstrations.
AutoCompressor recursively compresses concatenated demonstrations into soft prompts step-by-step.
However, AutoCompressor still underperforms SelfCP.
We attribute this to the information lost due to recursive compression in addressing long prompts.
Moreover, demonstrations filtered by LLMlingua generally underperform 0-shot in both XSUM and ARXIV with two backbones proving that the discrete tokens fail to guide LLM in few-shot settings.
We evaluate SelfCP on CoLA through closed-end evaluation, which measures the perplexity (PPL)~\footnote{\url{https://huggingface.co/docs/transformers/ perplexity.}} of candidate labels (acceptable/unacceptable) for the given sentence in the following template: \underline{[Sentence] Grammatically, the above sentence is \{acceptable/unacceptable\}}.
Labels with PPL closer to 1 will be judged as the prediction.
Notably, ICAE always returns ``acceptable'', resulting in a consent 30.9\% accuracy.

Finally, we evaluate SelfCP on selecting suitable demonstrations from MDB, which contains 5k randomly sampled demonstrations.
SelfCP further achieves performance gains by retrieving more powerful in-context demonstrations from MDB.
The result indicates that the compressed virtual token generated by SelfCP is good at measuring similarity among documents and then finding more favorable in-context demonstrations for ICL.

\begin{table*}
  \centering
    \caption{Case studies on the DUC dataset.
  The \textit{italic text} in Input represents the document to be condensed, while the \textbf{bold text} indicates the salient information derived from the compressed prompts.}
  \begin{adjustbox}{width={0.95\textwidth},totalheight={\textheight},keepaspectratio}
    \begin{tabular}{l|p{0.85\textwidth}}
    \bottomrule
    \multicolumn{2}{c}{DUC Case} \\
    \hline
    Input & According to the topic: Write an account of the sequence of events involving the Kennedy family... Summarize the following document: John F. Kennedy Jr. was a relatively new pilot... \textit{the plane could glide if the single engine failed </P><P> Kennedy had registered the plane, \textbf{a Piper Saratoga}...} \\
    \hline
    Reference & At approximately 9:40pm Friday, ..., when a \textbf{Piper Saratoga, piloted by John}, ... An at sea burial of all three was conducted ... \\
    \hline
    Vicuna & The sequence of events involving the Kennedy family following the plane crash that killed John F. Kennedy Jr., his wife, Carolyn Bessette Kennedy...\\
    \hline
    LLMLingua & The sequence of events involving the Kennedy family during and following the plane crash that killed John F. Kennedy Jr., his wife, Carolyn Bessette Kennedy...\\
    \hline
    AutoCompressor & </P><P> The plane was a twin-engine, single-seat Cessna Citation500, which costs about \$1.5 million, according to the manufacturer.</P><P>... \\
    \hline
    SelfCP & On July 17, 1999, John F. Kennedy Jr., his wife, ... \textbf{The plane was a Piper Saratoga, a twin-engine aircraft that was owned by Kennedy}... \\

    \toprule
    \end{tabular}%
    \end{adjustbox}
  \label{tab:casestudy}%
\end{table*}%

\begin{figure*}[t]
\centering
\includegraphics[width=0.95\textwidth]{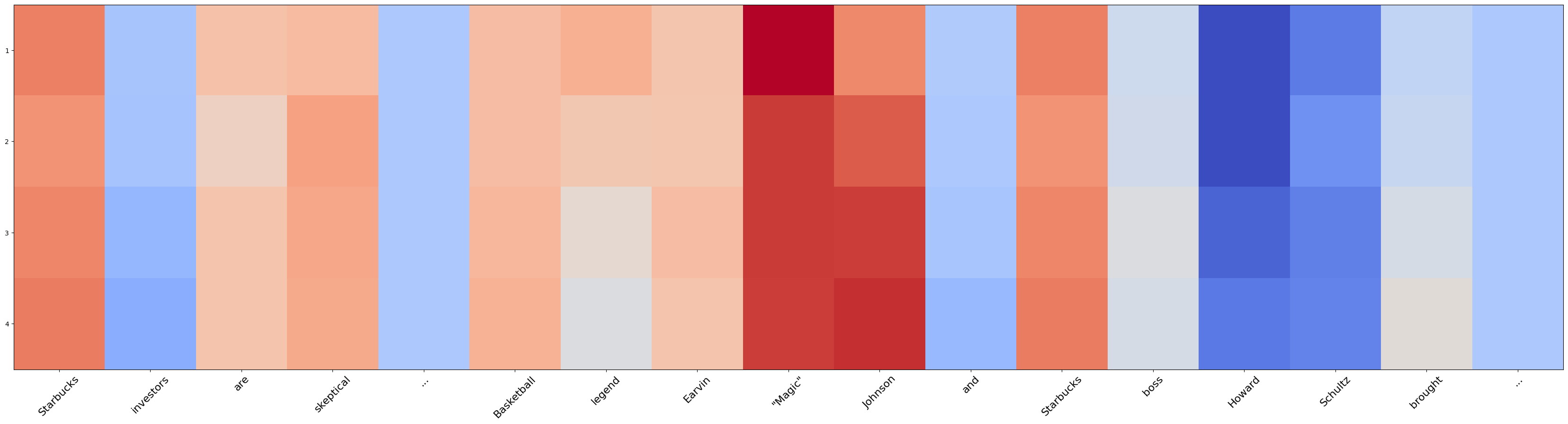}
\caption{The visualization result between condensed prompt and their virtual tokens.
To clarify, we sample 4 virtual tokens and select some representative actual tokens.}
\label{fig:case}
\end{figure*}

\subsection{Case Study}
We demonstrate a case study on DUC to provide an intuitive comparison among SelfCP based on Vicuna, direct transaction, AutoCompressor, and LLMLingua in Table~\ref{tab:casestudy}.
This case describes that a plane crashed into the Atlantic Ocean, while the details of the plane are over-length and condensed.
Vicuna can't generate satisfying summaries as some salient parts of the documents are truncated.
By contrast, SelfCP successfully restores the important information from the compact virtual tokens, such as the aircraft type ``Piper Saratoga''.

Furthermore, we depict the other scenario by gauging the similarity between actual and their virtual tokens, illustrated in Figure~\ref{fig:case}.
Warmer colors signify a greater degree of similarity.
The origin document describes the Stuckbarks and its cooperation with Magic Johnson, with the information about ``Magic Johnson'' compressed.
However, SelfCP recovers this information in the generated response.
It is plausible that the virtual tokens effectively absorb pertinent information, resulting in a comparatively higher similarity to these tokens.

\section{Analysis}
\begin{figure}[h]
\centering
\includegraphics[width=0.6\columnwidth]{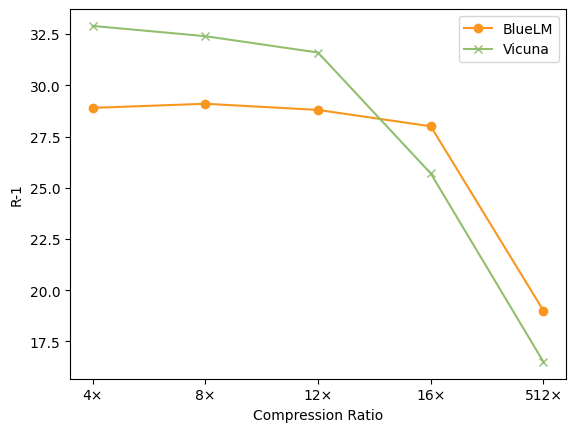}
\caption{The sensitivity analysis of compression ratio.}
\label{fig:ratio}
\end{figure}
\subsection{Compression Ratio}
The compression ratio is randomly sampled from 2 to 16 during the training of SelfCP.
We mix up 2,000 instances from the in-domain validation set, 1,000 for XSUM, and 1,000 for CICERO to select the compression ratio.
Specifically, SelfCP conducts conditional compression that compresses the latter cut-off part while keeping the former uncompressed.
Therefore, we can measure the information quality of the same content with different compression ratios by ROUGE-1 since it is more sensitive to token-level differences.

For both BlueLM and Vicuna, the performance is relative smoothing when the compression ratio changes from $4\times$ to $12\times$.
However, when it comes to $16\times$, a significant drop occurs in Vicuna, with relatively large performance degradation occurring in BlueLM as well compared to the previous ratios.
Therefore, we set the compression ratio to 12 by default and apply this ratio to all experiments.
Additionally, in our experiment setting, the window limitation is 512, and the $512\times$ compression ratio is equal to compressing anything to a single virtual token.

\begin{table}
\centering
\caption{The efficiency of SelfCP with the backbone of Vicuna. 1k stands for the extended 1k window limitation, while others have a 512 window limitation (ignoring the length of memory tokens).}
\begin{tabular}{l>{\centering\arraybackslash}p{1.8cm} >{\centering\arraybackslash}p{1.8cm}>{\centering\arraybackslash}p{1.8cm}}
\bottomrule
Method & GPUHours & TFLOPs & TMACs\\
\hline
Vicuna & 1.5 & 86,20 & 4,309\\
Vicuna-1k & 1.9 &31,664 & 15,832  \\

SelfCP& 1.6 & 22,437 & 11,218\\
\toprule
\end{tabular}
\label{tab:efficiency}
\end{table}

\begin{figure}
    \centering
    \subfigure{
        \includegraphics[width=0.4\textwidth]{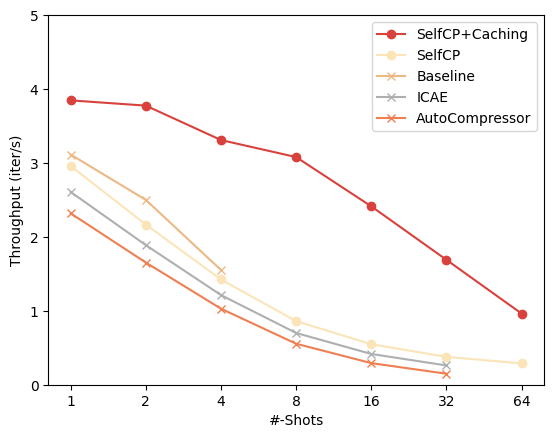}
        \label{fig:throughput}
    }
    \hfill
    \subfigure{
        \includegraphics[width=0.4\textwidth]{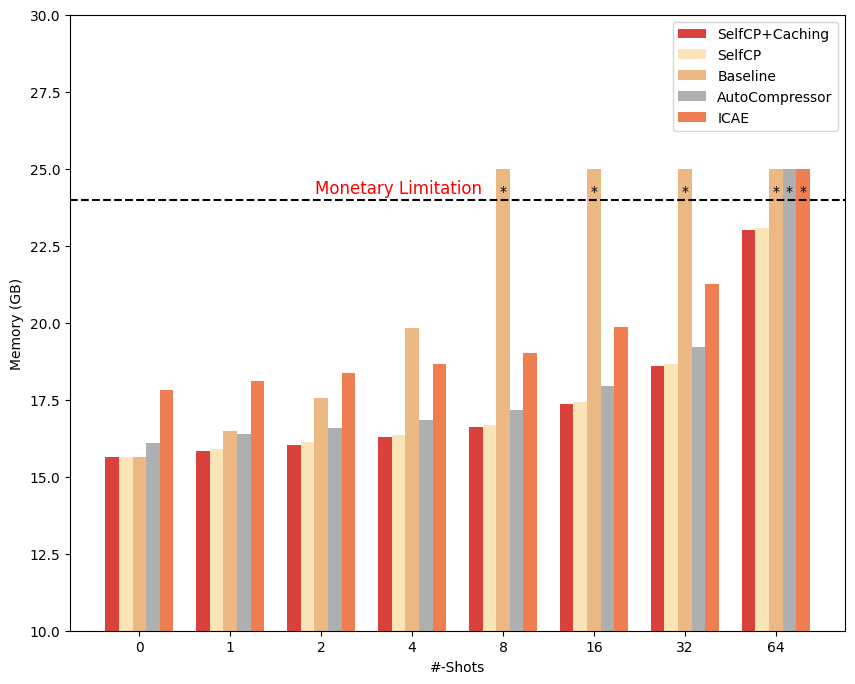}
        \label{fig:memory}
    }
    \caption{Efficiency comparison on throughputs (left) and memory costs (right) in the CoLA dataset. The upper limit of memory is 24G, and the excess part is marked with *, corresponding to the breaking point in the line chart.}
    \label{fig:efficiency}
\end{figure}

\subsection{Efficiency Analysis}
In SelfCP, we incorporate an additional 17M trainable parameters into the 7b backbone, accounting for an approximate increase of 0.24\%.

To quantify the efficiency difference brought by the projection layer, we mainly focus on SelfCP built on Vicuna because BlueLM has a comparable parameter size and model architecture. 
We first report the GPU Hours, TFLOPs, and TMACs~\footnote{\url{https://github.com/MrYxJ/calculate-flops.pytorch.}} of SelfCP and Vicuna on a single NVIDIA A5000 GPU.
Specifically, we use 1000 random but legal number sequences of 1024 length as input ids, avoiding special tokens, and ask the model to always generate 128 tokens.
SelfCP compresses the later 512 tokens into 43 memory tokens (12 $\times$ compression), and the former 512 tokens along with memory tokens are fed into Vicuna to perform generation (555 tokens in total).
To maintain a consistent input length for generation, the former 555 tokens are directly fed into Vicuna.
In Table~\ref{tab:efficiency}, SelfCP achieves nearly three times the TFLOPs and TMACs than the naive backbone to exchange double readable windows due to the additional compression and projection processes.
However, the forward propagation for compressing and the projection support parallel computing, and it only brings minimal GPU Hours increments in practice.
Notably, when allowing Vicuna to read the entire 1024 input ids without compression, the extra 512 tokens surge both computation and GPU Hours and overwhelm SelfCP during generation (referring to Vicuna-1k in Tab.~\ref{tab:efficiency}).

Except for the floats computation, in the ICL scenario, we illustrate the practical variations of throughput and memory costs with the number of demonstrations increasing among SelfCP and other compression methods in Figure~\ref{fig:efficiency}.
AutoCompressor exhibits poor inference throughputs due to recursively compression and ICAE performs sub-weakly in throughputs because of switching LoRA weights between the compressor and the generator.
Given a 24 GB memory allocation, Vicuna-7b only performs up to the 4-shot setting, while ICAE and AutoCompressor read up to 32 demonstrations, but SelfCP still works in the 64-shot setting.
SelfCP supports caching memory tokens to configure the Memory Demonstration Bank in advance for reusing.
SelfCP can respond to queries without compressing demonstration repeatedly in this case (referring to SelfCP + Caching), making inference more efficient.

\section{Conclusion and Further Work}
This paper proposes SelfCP, using a frozen LLM to compress long prompts into memory tokens.
In SelfCP, the LLM plays the role of the compressor to compress the prompt and the generator to respond to queries conditioned on the memory tokens and the rest uncompressed prompt.
SelfCP only contains 17M trainable parameters due to the frozen backbone and allows for adaptation across various backbones.
We conduct extensive in- and out-domain experiments, covering situations of ICL and over-length prompts and we analyze the efficiency of SelfCP.
The results show that the generated memory tokens can effectively substitute the 12$\times$ longer actual over-limit prompt.

We believe there is much improvement room for SelfCP.
On the one hand, we will scale the backbone of SelfCP to larger and higher-performance LLMs in various domains.
On the other hand, our intention involves incorporating compression as one of the fundamental pre-training goals of LLMs, expecting to enhance their compression ability further.

\printcredits

\bibliographystyle{cas-model2-names}

\bibliography{cas-refs}

\begin{thebibliography}{44}
\expandafter\ifx\csname natexlab\endcsname\relax\def\natexlab#1{#1}\fi
\providecommand{\url}[1]{\texttt{#1}}
\providecommand{\href}[2]{#2}
\providecommand{\path}[1]{#1}
\providecommand{\DOIprefix}{doi:}
\providecommand{\ArXivprefix}{arXiv:}
\providecommand{\URLprefix}{URL: }
\providecommand{\Pubmedprefix}{pmid:}
\providecommand{\doi}[1]{\href{http://dx.doi.org/#1}{\path{#1}}}
\providecommand{\Pubmed}[1]{\href{pmid:#1}{\path{#1}}}
\providecommand{\bibinfo}[2]{#2}
\ifx\xfnm\relax \def\xfnm[#1]{\unskip,\space#1}\fi
\bibitem[{Beltagy et~al.(2020)Beltagy, Peters and Cohan}]{beltagy2020longformer}
\bibinfo{author}{Beltagy, I.}, \bibinfo{author}{Peters, M.E.}, \bibinfo{author}{Cohan, A.}, \bibinfo{year}{2020}.
\newblock \bibinfo{title}{Longformer: The long-document transformer}.
\newblock \bibinfo{journal}{arXiv preprint arXiv:2004.05150} .
\bibitem[{Bertsch et~al.(2023)Bertsch, Alon, Neubig and Gormley}]{bertsch2023unlimiformer}
\bibinfo{author}{Bertsch, A.}, \bibinfo{author}{Alon, U.}, \bibinfo{author}{Neubig, G.}, \bibinfo{author}{Gormley, M.R.}, \bibinfo{year}{2023}.
\newblock \bibinfo{title}{Unlimiformer: Long-range transformers with unlimited length input}.
\newblock \bibinfo{journal}{arXiv preprint arXiv:2305.01625} .
\bibitem[{Bolya et~al.(2022)Bolya, Fu, Dai, Zhang, Feichtenhofer and Hoffman}]{bolya2022token}
\bibinfo{author}{Bolya, D.}, \bibinfo{author}{Fu, C.Y.}, \bibinfo{author}{Dai, X.}, \bibinfo{author}{Zhang, P.}, \bibinfo{author}{Feichtenhofer, C.}, \bibinfo{author}{Hoffman, J.}, \bibinfo{year}{2022}.
\newblock \bibinfo{title}{Token merging: Your vit but faster}.
\newblock \bibinfo{journal}{arXiv preprint arXiv:2210.09461} .
\bibitem[{Bulatov et~al.(2022)Bulatov, Kuratov and Burtsev}]{bulatov2022recurrent}
\bibinfo{author}{Bulatov, A.}, \bibinfo{author}{Kuratov, Y.}, \bibinfo{author}{Burtsev, M.}, \bibinfo{year}{2022}.
\newblock \bibinfo{title}{Recurrent memory transformer}.
\newblock \bibinfo{journal}{Advances in Neural Information Processing Systems} \bibinfo{volume}{35}, \bibinfo{pages}{11079--11091}.
\bibitem[{Bulatov et~al.(2023)Bulatov, Kuratov, Kapushev and Burtsev}]{bulatov2023scaling}
\bibinfo{author}{Bulatov, A.}, \bibinfo{author}{Kuratov, Y.}, \bibinfo{author}{Kapushev, Y.}, \bibinfo{author}{Burtsev, M.S.}, \bibinfo{year}{2023}.
\newblock \bibinfo{title}{Scaling transformer to 1m tokens and beyond with rmt}.
\newblock \bibinfo{journal}{arXiv preprint arXiv:2304.11062} .
\bibitem[{Chevalier et~al.(2023)Chevalier, Wettig, Ajith and Chen}]{chevalier2023adapting}
\bibinfo{author}{Chevalier, A.}, \bibinfo{author}{Wettig, A.}, \bibinfo{author}{Ajith, A.}, \bibinfo{author}{Chen, D.}, \bibinfo{year}{2023}.
\newblock \bibinfo{title}{Adapting language models to compress contexts}.
\newblock \bibinfo{journal}{arXiv preprint arXiv:2305.14788} .
\bibitem[{Child et~al.(2019)Child, Gray, Radford and Sutskever}]{child2019generating}
\bibinfo{author}{Child, R.}, \bibinfo{author}{Gray, S.}, \bibinfo{author}{Radford, A.}, \bibinfo{author}{Sutskever, I.}, \bibinfo{year}{2019}.
\newblock \bibinfo{title}{Generating long sequences with sparse transformers}.
\newblock \bibinfo{journal}{arXiv preprint arXiv:1904.10509} .
\bibitem[{Choromanski et~al.(2020)Choromanski, Likhosherstov, Dohan, Song, Gane, Sarlos, Hawkins, Davis, Mohiuddin, Kaiser et~al.}]{choromanski2020rethinking}
\bibinfo{author}{Choromanski, K.}, \bibinfo{author}{Likhosherstov, V.}, \bibinfo{author}{Dohan, D.}, \bibinfo{author}{Song, X.}, \bibinfo{author}{Gane, A.}, \bibinfo{author}{Sarlos, T.}, \bibinfo{author}{Hawkins, P.}, \bibinfo{author}{Davis, J.}, \bibinfo{author}{Mohiuddin, A.}, \bibinfo{author}{Kaiser, L.}, et~al., \bibinfo{year}{2020}.
\newblock \bibinfo{title}{Rethinking attention with performers}.
\newblock \bibinfo{journal}{arXiv preprint arXiv:2009.14794} .
\bibitem[{Cohan et~al.(2018)Cohan, Dernoncourt, Kim, Bui, Kim, Chang and Goharian}]{cohan2018discourse}
\bibinfo{author}{Cohan, A.}, \bibinfo{author}{Dernoncourt, F.}, \bibinfo{author}{Kim, D.S.}, \bibinfo{author}{Bui, T.}, \bibinfo{author}{Kim, S.}, \bibinfo{author}{Chang, W.}, \bibinfo{author}{Goharian, N.}, \bibinfo{year}{2018}.
\newblock \bibinfo{title}{A discourse-aware attention model for abstractive summarization of long documents}.
\newblock \bibinfo{journal}{arXiv preprint arXiv:1804.05685} .
\bibitem[{Copeck et~al.(2006)Copeck, Inkpen, Kazantseva, Kennedy, Kipp, Nastase and Szpakowicz}]{copeck2006leveraging}
\bibinfo{author}{Copeck, T.}, \bibinfo{author}{Inkpen, D.}, \bibinfo{author}{Kazantseva, A.}, \bibinfo{author}{Kennedy, A.}, \bibinfo{author}{Kipp, D.}, \bibinfo{author}{Nastase, V.}, \bibinfo{author}{Szpakowicz, S.}, \bibinfo{year}{2006}.
\newblock \bibinfo{title}{Leveraging duc}, in: \bibinfo{booktitle}{proceedings of DUC}.
\bibitem[{Dai et~al.(2019)Dai, Yang, Yang, Carbonell, Le and Salakhutdinov}]{dai2019transformer}
\bibinfo{author}{Dai, Z.}, \bibinfo{author}{Yang, Z.}, \bibinfo{author}{Yang, Y.}, \bibinfo{author}{Carbonell, J.}, \bibinfo{author}{Le, Q.V.}, \bibinfo{author}{Salakhutdinov, R.}, \bibinfo{year}{2019}.
\newblock \bibinfo{title}{Transformer-xl: Attentive language models beyond a fixed-length context}.
\newblock \bibinfo{journal}{arXiv preprint arXiv:1901.02860} .
\bibitem[{Ding et~al.(2023)Ding, Ma, Dong, Zhang, Huang, Wang and Wei}]{ding2023longnet}
\bibinfo{author}{Ding, J.}, \bibinfo{author}{Ma, S.}, \bibinfo{author}{Dong, L.}, \bibinfo{author}{Zhang, X.}, \bibinfo{author}{Huang, S.}, \bibinfo{author}{Wang, W.}, \bibinfo{author}{Wei, F.}, \bibinfo{year}{2023}.
\newblock \bibinfo{title}{Longnet: Scaling transformers to 1,000,000,000 tokens}.
\newblock \bibinfo{journal}{arXiv preprint arXiv:2307.02486} .
\bibitem[{Dosovitskiy et~al.(2020)Dosovitskiy, Beyer, Kolesnikov, Weissenborn, Zhai, Unterthiner, Dehghani, Minderer, Heigold, Gelly et~al.}]{dosovitskiy2020image}
\bibinfo{author}{Dosovitskiy, A.}, \bibinfo{author}{Beyer, L.}, \bibinfo{author}{Kolesnikov, A.}, \bibinfo{author}{Weissenborn, D.}, \bibinfo{author}{Zhai, X.}, \bibinfo{author}{Unterthiner, T.}, \bibinfo{author}{Dehghani, M.}, \bibinfo{author}{Minderer, M.}, \bibinfo{author}{Heigold, G.}, \bibinfo{author}{Gelly, S.}, et~al., \bibinfo{year}{2020}.
\newblock \bibinfo{title}{An image is worth 16x16 words: Transformers for image recognition at scale}.
\newblock \bibinfo{journal}{arXiv preprint arXiv:2010.11929} .
\bibitem[{Ge et~al.(2023)Ge, Hu, Wang, Chen and Wei}]{ge2023context}
\bibinfo{author}{Ge, T.}, \bibinfo{author}{Hu, J.}, \bibinfo{author}{Wang, X.}, \bibinfo{author}{Chen, S.Q.}, \bibinfo{author}{Wei, F.}, \bibinfo{year}{2023}.
\newblock \bibinfo{title}{In-context autoencoder for context compression in a large language model}.
\newblock \bibinfo{journal}{arXiv preprint arXiv:2307.06945} .
\bibitem[{Ghosal et~al.(2022)Ghosal, Shen, Majumder, Mihalcea and Poria}]{ghosal2022cicero}
\bibinfo{author}{Ghosal, D.}, \bibinfo{author}{Shen, S.}, \bibinfo{author}{Majumder, N.}, \bibinfo{author}{Mihalcea, R.}, \bibinfo{author}{Poria, S.}, \bibinfo{year}{2022}.
\newblock \bibinfo{title}{Cicero: A dataset for contextualized commonsense inference in dialogues}.
\newblock \bibinfo{journal}{arXiv preprint arXiv:2203.13926} .
\bibitem[{Jiang et~al.(2023)Jiang, Wu, Lin, Yang and Qiu}]{jiang2023llmlingua}
\bibinfo{author}{Jiang, H.}, \bibinfo{author}{Wu, Q.}, \bibinfo{author}{Lin, C.Y.}, \bibinfo{author}{Yang, Y.}, \bibinfo{author}{Qiu, L.}, \bibinfo{year}{2023}.
\newblock \bibinfo{title}{Llmlingua: Compressing prompts for accelerated inference of large language models}.
\newblock \bibinfo{journal}{arXiv preprint arXiv:2310.05736} .
\bibitem[{Katharopoulos et~al.(2020)Katharopoulos, Vyas, Pappas and Fleuret}]{katharopoulos2020transformers}
\bibinfo{author}{Katharopoulos, A.}, \bibinfo{author}{Vyas, A.}, \bibinfo{author}{Pappas, N.}, \bibinfo{author}{Fleuret, F.}, \bibinfo{year}{2020}.
\newblock \bibinfo{title}{Transformers are rnns: Fast autoregressive transformers with linear attention}, in: \bibinfo{booktitle}{International conference on machine learning}, \bibinfo{organization}{PMLR}. pp. \bibinfo{pages}{5156--5165}.
\bibitem[{Kim et~al.(2022)Kim, Shen, Thorsley, Gholami, Kwon, Hassoun and Keutzer}]{kim2022learned}
\bibinfo{author}{Kim, S.}, \bibinfo{author}{Shen, S.}, \bibinfo{author}{Thorsley, D.}, \bibinfo{author}{Gholami, A.}, \bibinfo{author}{Kwon, W.}, \bibinfo{author}{Hassoun, J.}, \bibinfo{author}{Keutzer, K.}, \bibinfo{year}{2022}.
\newblock \bibinfo{title}{Learned token pruning for transformers}, in: \bibinfo{booktitle}{Proceedings of the 28th ACM SIGKDD Conference on Knowledge Discovery and Data Mining}, pp. \bibinfo{pages}{784--794}.
\bibitem[{Lee-Thorp et~al.(2021)Lee-Thorp, Ainslie, Eckstein and Ontanon}]{lee2021fnet}
\bibinfo{author}{Lee-Thorp, J.}, \bibinfo{author}{Ainslie, J.}, \bibinfo{author}{Eckstein, I.}, \bibinfo{author}{Ontanon, S.}, \bibinfo{year}{2021}.
\newblock \bibinfo{title}{Fnet: Mixing tokens with fourier transforms}.
\newblock \bibinfo{journal}{arXiv preprint arXiv:2105.03824} .
\bibitem[{Li et~al.(2023a)Li, Li, Savarese and Hoi}]{li2023blip}
\bibinfo{author}{Li, J.}, \bibinfo{author}{Li, D.}, \bibinfo{author}{Savarese, S.}, \bibinfo{author}{Hoi, S.}, \bibinfo{year}{2023}a.
\newblock \bibinfo{title}{Blip-2: Bootstrapping language-image pre-training with frozen image encoders and large language models}.
\newblock \bibinfo{journal}{arXiv preprint arXiv:2301.12597} .
\bibitem[{Li and Liang(2021)}]{li2021prefix}
\bibinfo{author}{Li, X.L.}, \bibinfo{author}{Liang, P.}, \bibinfo{year}{2021}.
\newblock \bibinfo{title}{Prefix-tuning: Optimizing continuous prompts for generation}.
\newblock \bibinfo{journal}{arXiv preprint arXiv:2101.00190} .
\bibitem[{Li(2023)}]{li2023unlocking}
\bibinfo{author}{Li, Y.}, \bibinfo{year}{2023}.
\newblock \bibinfo{title}{Unlocking context constraints of llms: Enhancing context efficiency of llms with self-information-based content filtering}.
\newblock \bibinfo{journal}{arXiv preprint arXiv:2304.12102} .
\bibitem[{Li et~al.(2023b)Li, Yang, Wang, Wei and Li}]{li2023generative}
\bibinfo{author}{Li, Y.}, \bibinfo{author}{Yang, N.}, \bibinfo{author}{Wang, L.}, \bibinfo{author}{Wei, F.}, \bibinfo{author}{Li, W.}, \bibinfo{year}{2023}b.
\newblock \bibinfo{title}{Generative retrieval for conversational question answering}.
\newblock \bibinfo{journal}{Information Processing \& Management} \bibinfo{volume}{60}, \bibinfo{pages}{103475}.
\bibitem[{Lin(2004)}]{lin2004rouge}
\bibinfo{author}{Lin, C.Y.}, \bibinfo{year}{2004}.
\newblock \bibinfo{title}{Rouge: A package for automatic evaluation of summaries}, in: \bibinfo{booktitle}{Text summarization branches out}, pp. \bibinfo{pages}{74--81}.
\bibitem[{Liu et~al.(2024)Liu, Lin, Hewitt, Paranjape, Bevilacqua, Petroni and Liang}]{liu2024lost}
\bibinfo{author}{Liu, N.F.}, \bibinfo{author}{Lin, K.}, \bibinfo{author}{Hewitt, J.}, \bibinfo{author}{Paranjape, A.}, \bibinfo{author}{Bevilacqua, M.}, \bibinfo{author}{Petroni, F.}, \bibinfo{author}{Liang, P.}, \bibinfo{year}{2024}.
\newblock \bibinfo{title}{Lost in the middle: How language models use long contexts}.
\newblock \bibinfo{journal}{Transactions of the Association for Computational Linguistics} \bibinfo{volume}{12}, \bibinfo{pages}{157--173}.
\bibitem[{Liu et~al.(2022)Liu, Ji, Fu, Tam, Du, Yang and Tang}]{liu2022p}
\bibinfo{author}{Liu, X.}, \bibinfo{author}{Ji, K.}, \bibinfo{author}{Fu, Y.}, \bibinfo{author}{Tam, W.}, \bibinfo{author}{Du, Z.}, \bibinfo{author}{Yang, Z.}, \bibinfo{author}{Tang, J.}, \bibinfo{year}{2022}.
\newblock \bibinfo{title}{P-tuning: Prompt tuning can be comparable to fine-tuning across scales and tasks}, in: \bibinfo{booktitle}{Proceedings of the 60th Annual Meeting of the Association for Computational Linguistics (Volume 2: Short Papers)}, pp. \bibinfo{pages}{61--68}.
\bibitem[{Mu et~al.(2023)Mu, Li and Goodman}]{mu2023learning}
\bibinfo{author}{Mu, J.}, \bibinfo{author}{Li, X.L.}, \bibinfo{author}{Goodman, N.}, \bibinfo{year}{2023}.
\newblock \bibinfo{title}{Learning to compress prompts with gist tokens}.
\newblock \bibinfo{journal}{arXiv preprint arXiv:2304.08467} .
\bibitem[{Narayan et~al.(2018)Narayan, Cohen and Lapata}]{Narayan2018DontGM}
\bibinfo{author}{Narayan, S.}, \bibinfo{author}{Cohen, S.B.}, \bibinfo{author}{Lapata, M.}, \bibinfo{year}{2018}.
\newblock \bibinfo{title}{Don't give me the details, just the summary! topic-aware convolutional neural networks for extreme summarization}.
\newblock \bibinfo{journal}{ArXiv} \bibinfo{volume}{abs/1808.08745}.
\bibitem[{Sun et~al.(2024)Sun, Yuan, Li, Cao and Li}]{sun2024dialogue}
\bibinfo{author}{Sun, S.}, \bibinfo{author}{Yuan, R.}, \bibinfo{author}{Li, W.}, \bibinfo{author}{Cao, Z.}, \bibinfo{author}{Li, S.}, \bibinfo{year}{2024}.
\newblock \bibinfo{title}{Dialogue acts enhanced extract--abstract framework for meeting summarization}.
\newblock \bibinfo{journal}{Information Processing \& Management} \bibinfo{volume}{61}, \bibinfo{pages}{103635}.
\bibitem[{Taori et~al.(2023)Taori, Gulrajani, Zhang, Dubois, Li, Guestrin, Liang and Hashimoto}]{taori2023stanford}
\bibinfo{author}{Taori, R.}, \bibinfo{author}{Gulrajani, I.}, \bibinfo{author}{Zhang, T.}, \bibinfo{author}{Dubois, Y.}, \bibinfo{author}{Li, X.}, \bibinfo{author}{Guestrin, C.}, \bibinfo{author}{Liang, P.}, \bibinfo{author}{Hashimoto, T.B.}, \bibinfo{year}{2023}.
\newblock \bibinfo{title}{Stanford alpaca: An instruction-following llama model}.
\bibitem[{Team(2023)}]{2023bluelm}
\bibinfo{author}{Team, B.}, \bibinfo{year}{2023}.
\newblock \bibinfo{title}{Bluelm: An open multilingual 7b language model}.
\newblock \bibinfo{howpublished}{\url{https://github.com/vivo-ai-lab/BlueLM}}.
\bibitem[{Touvron et~al.(2023)Touvron, Lavril, Izacard, Martinet, Lachaux, Lacroix, Rozi{\`e}re, Goyal, Hambro, Azhar et~al.}]{touvron2023llama}
\bibinfo{author}{Touvron, H.}, \bibinfo{author}{Lavril, T.}, \bibinfo{author}{Izacard, G.}, \bibinfo{author}{Martinet, X.}, \bibinfo{author}{Lachaux, M.A.}, \bibinfo{author}{Lacroix, T.}, \bibinfo{author}{Rozi{\`e}re, B.}, \bibinfo{author}{Goyal, N.}, \bibinfo{author}{Hambro, E.}, \bibinfo{author}{Azhar, F.}, et~al., \bibinfo{year}{2023}.
\newblock \bibinfo{title}{Llama: Open and efficient foundation language models}.
\newblock \bibinfo{journal}{arXiv preprint arXiv:2302.13971} .
\bibitem[{Wang et~al.(2023a)Wang, Liang, Meng, Shi, Li, Xu, Qu and Zhou}]{wang12023chatgpt}
\bibinfo{author}{Wang, J.}, \bibinfo{author}{Liang, Y.}, \bibinfo{author}{Meng, F.}, \bibinfo{author}{Shi, H.}, \bibinfo{author}{Li, Z.}, \bibinfo{author}{Xu, J.}, \bibinfo{author}{Qu, J.}, \bibinfo{author}{Zhou, J.}, \bibinfo{year}{2023}a.
\newblock \bibinfo{title}{Is chatgpt a good nlg evaluator? a preliminary study}.
\newblock \bibinfo{journal}{arXiv preprint arXiv:2303.04048} .
\bibitem[{Wang et~al.(2024)Wang, Zhao, Ji, Jiang, Li, Hu and Lu}]{wang2024dialogue}
\bibinfo{author}{Wang, L.}, \bibinfo{author}{Zhao, M.}, \bibinfo{author}{Ji, H.}, \bibinfo{author}{Jiang, Z.}, \bibinfo{author}{Li, R.}, \bibinfo{author}{Hu, Z.}, \bibinfo{author}{Lu, X.}, \bibinfo{year}{2024}.
\newblock \bibinfo{title}{Dialogue summarization enhanced response generation for multi-domain task-oriented dialogue systems}.
\newblock \bibinfo{journal}{Information Processing \& Management} \bibinfo{volume}{61}, \bibinfo{pages}{103668}.
\bibitem[{Wang et~al.(2022)Wang, Mishra, Alipoormolabashi, Kordi, Mirzaei, Arunkumar, Ashok, Dhanasekaran, Naik, Stap et~al.}]{wang2022super}
\bibinfo{author}{Wang, Y.}, \bibinfo{author}{Mishra, S.}, \bibinfo{author}{Alipoormolabashi, P.}, \bibinfo{author}{Kordi, Y.}, \bibinfo{author}{Mirzaei, A.}, \bibinfo{author}{Arunkumar, A.}, \bibinfo{author}{Ashok, A.}, \bibinfo{author}{Dhanasekaran, A.S.}, \bibinfo{author}{Naik, A.}, \bibinfo{author}{Stap, D.}, et~al., \bibinfo{year}{2022}.
\newblock \bibinfo{title}{Super-naturalinstructions: Generalization via declarative instructions on 1600+ nlp tasks}.
\newblock \bibinfo{journal}{arXiv preprint arXiv:2204.07705} .
\bibitem[{Wang et~al.(2023b)Wang, Xie, Ding, Feng and Xia}]{wang2023chatgpt}
\bibinfo{author}{Wang, Z.}, \bibinfo{author}{Xie, Q.}, \bibinfo{author}{Ding, Z.}, \bibinfo{author}{Feng, Y.}, \bibinfo{author}{Xia, R.}, \bibinfo{year}{2023}b.
\newblock \bibinfo{title}{Is chatgpt a good sentiment analyzer? a preliminary study}.
\newblock \bibinfo{journal}{arXiv preprint arXiv:2304.04339} .
\bibitem[{Wei et~al.(2023)Wei, Cui, Cheng, Wang, Zhang, Huang, Xie, Xu, Chen, Zhang et~al.}]{wei2023zero}
\bibinfo{author}{Wei, X.}, \bibinfo{author}{Cui, X.}, \bibinfo{author}{Cheng, N.}, \bibinfo{author}{Wang, X.}, \bibinfo{author}{Zhang, X.}, \bibinfo{author}{Huang, S.}, \bibinfo{author}{Xie, P.}, \bibinfo{author}{Xu, J.}, \bibinfo{author}{Chen, Y.}, \bibinfo{author}{Zhang, M.}, et~al., \bibinfo{year}{2023}.
\newblock \bibinfo{title}{Zero-shot information extraction via chatting with chatgpt}.
\newblock \bibinfo{journal}{arXiv preprint arXiv:2302.10205} .
\bibitem[{Wingate et~al.(2022)Wingate, Shoeybi and Sorensen}]{wingate2022prompt}
\bibinfo{author}{Wingate, D.}, \bibinfo{author}{Shoeybi, M.}, \bibinfo{author}{Sorensen, T.}, \bibinfo{year}{2022}.
\newblock \bibinfo{title}{Prompt compression and contrastive conditioning for controllability and toxicity reduction in language models}.
\newblock \bibinfo{journal}{arXiv preprint arXiv:2210.03162} .
\bibitem[{Wu et~al.(2022)Wu, Rabe, Hutchins and Szegedy}]{wu2022memorizing}
\bibinfo{author}{Wu, Y.}, \bibinfo{author}{Rabe, M.N.}, \bibinfo{author}{Hutchins, D.}, \bibinfo{author}{Szegedy, C.}, \bibinfo{year}{2022}.
\newblock \bibinfo{title}{Memorizing transformers}.
\newblock \bibinfo{journal}{arXiv preprint arXiv:2203.08913} .
\bibitem[{Yang et~al.(2023)Yang, Li, Zhang, Chen and Cheng}]{yang2023exploring}
\bibinfo{author}{Yang, X.}, \bibinfo{author}{Li, Y.}, \bibinfo{author}{Zhang, X.}, \bibinfo{author}{Chen, H.}, \bibinfo{author}{Cheng, W.}, \bibinfo{year}{2023}.
\newblock \bibinfo{title}{Exploring the limits of chatgpt for query or aspect-based text summarization}.
\newblock \bibinfo{journal}{arXiv preprint arXiv:2302.08081} .
\bibitem[{Zhang et~al.(2022)Zhang, Roller, Goyal, Artetxe, Chen, Chen, Dewan, Diab, Li, Lin et~al.}]{zhang2022opt}
\bibinfo{author}{Zhang, S.}, \bibinfo{author}{Roller, S.}, \bibinfo{author}{Goyal, N.}, \bibinfo{author}{Artetxe, M.}, \bibinfo{author}{Chen, M.}, \bibinfo{author}{Chen, S.}, \bibinfo{author}{Dewan, C.}, \bibinfo{author}{Diab, M.}, \bibinfo{author}{Li, X.}, \bibinfo{author}{Lin, X.V.}, et~al., \bibinfo{year}{2022}.
\newblock \bibinfo{title}{Opt: Open pre-trained transformer language models}.
\newblock \bibinfo{journal}{arXiv preprint arXiv:2205.01068} .
\bibitem[{Zheng et~al.(2023)Zheng, Chiang, Sheng, Zhuang, Wu, Zhuang, Lin, Li, Li, Xing et~al.}]{zheng2023judging}
\bibinfo{author}{Zheng, L.}, \bibinfo{author}{Chiang, W.L.}, \bibinfo{author}{Sheng, Y.}, \bibinfo{author}{Zhuang, S.}, \bibinfo{author}{Wu, Z.}, \bibinfo{author}{Zhuang, Y.}, \bibinfo{author}{Lin, Z.}, \bibinfo{author}{Li, Z.}, \bibinfo{author}{Li, D.}, \bibinfo{author}{Xing, E.}, et~al., \bibinfo{year}{2023}.
\newblock \bibinfo{title}{Judging llm-as-a-judge with mt-bench and chatbot arena}.
\newblock \bibinfo{journal}{arXiv preprint arXiv:2306.05685} .
\bibitem[{Zheng et~al.(2022)Zheng, Wang and Kong}]{zheng2022linear}
\bibinfo{author}{Zheng, L.}, \bibinfo{author}{Wang, C.}, \bibinfo{author}{Kong, L.}, \bibinfo{year}{2022}.
\newblock \bibinfo{title}{Linear complexity randomized self-attention mechanism}, in: \bibinfo{booktitle}{International conference on machine learning}, \bibinfo{organization}{PMLR}. pp. \bibinfo{pages}{27011--27041}.
\bibitem[{Zhong et~al.(2022)Zhong, Lei and Chen}]{zhong2022training}
\bibinfo{author}{Zhong, Z.}, \bibinfo{author}{Lei, T.}, \bibinfo{author}{Chen, D.}, \bibinfo{year}{2022}.
\newblock \bibinfo{title}{Training language models with memory augmentation}.
\newblock \bibinfo{journal}{arXiv preprint arXiv:2205.12674} .

\end{thebibliography}

\appendix
\newpage
\section{Dataset}
We briefly introduce each dataset involved in this paper as follows:
\begin{table}[h]
    \centering
        \caption{Used instructions in involved datasets.}
    \begin{tabular}{c|l}
    \bottomrule
    
        XSUM & Summarize the following document: [DOC]\\
        \hline
        CICERO & [DOC] According to the above text, answer the query: [Query]\\
        \hline
        ARXIV & Summarize the following document: [DOC]\\
        \hline
        DUC & According to the topic: [Query]  Summarize the following documents: [DOC1][DOC2]...\\
        \hline
        CLCV & Generating the judgment according to the following indictment: [DOC]\\
        \hline
        CoLA &  [Sentence] Grammatically, the above sentence is \{acceptable/unacceptable\}\\
    \toprule
    \end{tabular}

    \label{tab:my_label}
\end{table}
\paragraph{XSUM}
XSUM \cite{Narayan2018DontGM} is a popular abstractive summarization dataset.
The documents in the XSUM dataset come from articles in various fields, including news stories, encyclopedia entries, forum posts, and more.
Each instance contains a document and its summary.


\paragraph{CICERO}
CICERO \cite{ghosal2022cicero} is a generative question-answering dataset.
Each instance contains a personal dialogue, a target speech, a query, and a human-written answer.
Language models need to answer the query conditioned on the target speech.

\paragraph{SUPER-NI}
SUPER-NI \cite{wang2022super} is a comprehensive instruction dataset including thousands of common natural language processing tasks.
Each instance contains an instruction and a response.
We introduce this dataset to make SelfCP able to respond to diverse queries.

\paragraph{DUC}
DUC 2007 \cite{copeck2006leveraging} is a popular Query Focused Summarization (QFS) dataset.
Each instance contains a topic and a few documents.

\paragraph{ARXIV}
ARXIV \cite{cohan2018discourse} is a summarization dataset collected from arXiv.
Each instance of this dataset contains the body of a paper and its abstract.
Therefore, ARXIV is relatively longer than normal datasets, in which documents contain 5.9k words on average.
\paragraph{CoLA}
The Corpus of Linguistic Acceptability (CoLA) is a popular linguistic acceptability dataset consisting of 10657 sentences from 23 linguistics publications, expertly annotated for acceptability (grammaticality) by their original authors.




\end{document}